\documentclass[10pt,twocolumn,letterpaper]{article}

\usepackage{iccv}              

\definecolor{iccvblue}{rgb}{0.21,0.49,0.74}
\usepackage[pagebackref,breaklinks,colorlinks,allcolors=iccvblue]{hyperref}
\usepackage[lined,ruled]{algorithm2e}
\usepackage{algorithmic}
\usepackage{setspace}
\usepackage{multirow}
\usepackage{stfloats}
\usepackage{graphicx}
\def\paperID{9514} 
\def\confName{ICCV}
\def\confYear{2025}
\title{Diffusion Transformer meets Multi-level Wavelet Spectrum \\ for Single Image Super-Resolution}

\author{
    Peng Du\textsuperscript{1},\hspace{0.5em}
    Hui Li\textsuperscript{1}\thanks{Corresponding author.},\hspace{0.5em}
    Han Xu\textsuperscript{1},\hspace{0.5em}
    Paul Barom Jeon\textsuperscript{2},\hspace{0.5em}
    Dongwook Lee\textsuperscript{2},\hspace{0.5em}
    Daehyun Ji\textsuperscript{2}, \\
    Ran Yang\textsuperscript{1},\hspace{0.5em}
    Feng Zhu\textsuperscript{1} \\
  \textsuperscript{1}Samsung R\&D Institute China Xi'an (SRCX) \\
  \textsuperscript{2}Samsung Electronics Co., LTD., South Korea \\
  \small
  \texttt{peng03.du, hui01.li, han.xu, paul.barom.jeon, dw12.lee, derek.ji,} \\
  \small
  \texttt{ran01.yang, f15.zhu@samsung.com}
}
\begin{document}
\pagestyle{plain} 
\maketitle
\begin{abstract}
    Discrete Wavelet Transform (DWT) has been widely explored to enhance the performance of image super-resolution (SR). Despite some DWT-based methods improving SR by capturing fine-grained frequency signals, most existing approaches neglect the interrelations among multi-scale frequency sub-bands, resulting in inconsistencies and unnatural artifacts in the reconstructed images. To address this challenge, we propose a Diffusion Transformer model based on image Wavelet spectra for SR (DTWSR). DTWSR incorporates the superiority of diffusion models and transformers to capture the interrelations among multi-scale frequency sub-bands, leading to a more consistence and realistic SR image. Specifically, we use a Multi-level Discrete Wavelet Transform (MDWT) to decompose images into wavelet spectra. A pyramid tokenization method is proposed which embeds the spectra into a sequence of tokens for transformer model, facilitating to capture features from both spatial and frequency domain. A dual-decoder is designed elaborately to handle the distinct variances in low-frequency (LF) and high-frequency (HF) sub-bands, without omitting their alignment in image generation. 
    Extensive experiments on multiple benchmark datasets demonstrate the effectiveness of our method, with high performance on both perception quality and fidelity. 
\end{abstract}    
\section{Introduction}
\label{sec:intro}

\begin{figure}
  \centering
  \begin{minipage}[c]{0.47\textwidth}
    \centering
    \includegraphics[width=\textwidth]{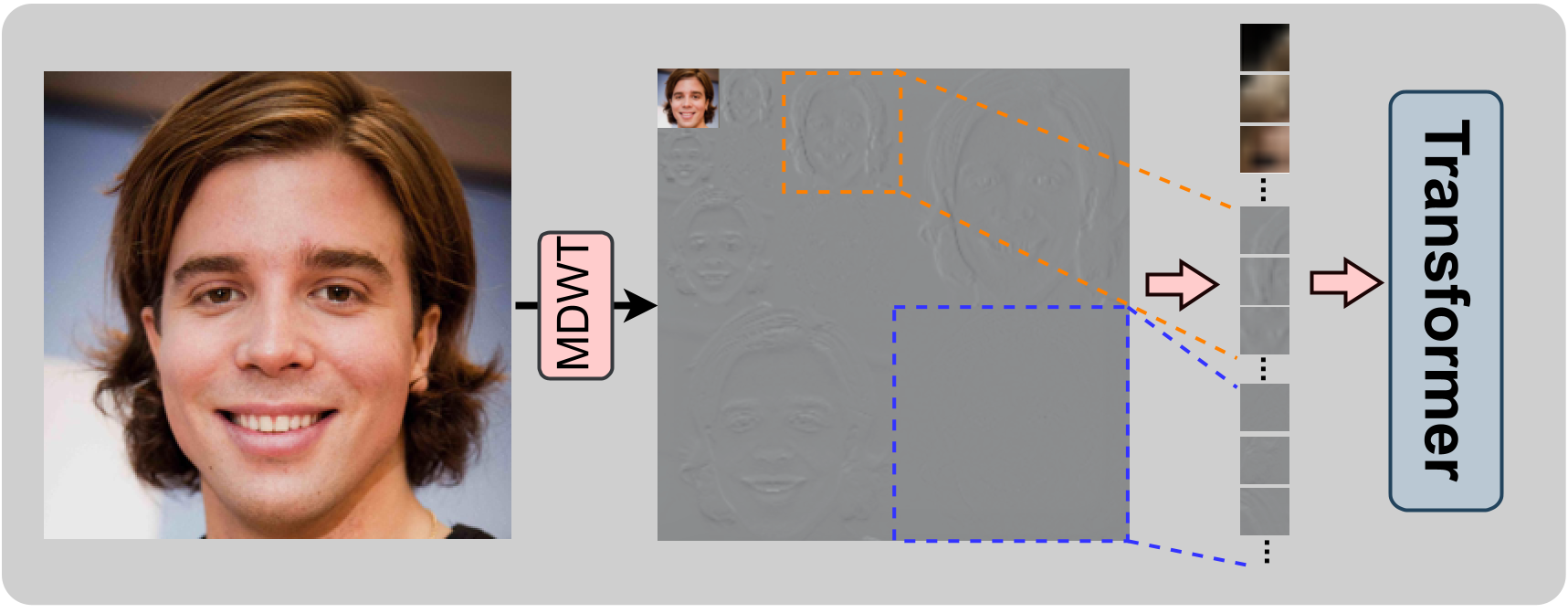}
    \subcaption{The pixel image is represented by multi-level wavelet spectra. Transform is explored to model the complex relations among the multi-scale frequencies.}
    \label{fig_mdwt_a}
  \end{minipage} 
  \\
  \begin{minipage}[c]{0.47\textwidth}
    \centering
    \begin{minipage}[t]{0.30\textwidth}
      \centering
      \includegraphics[width=\textwidth]{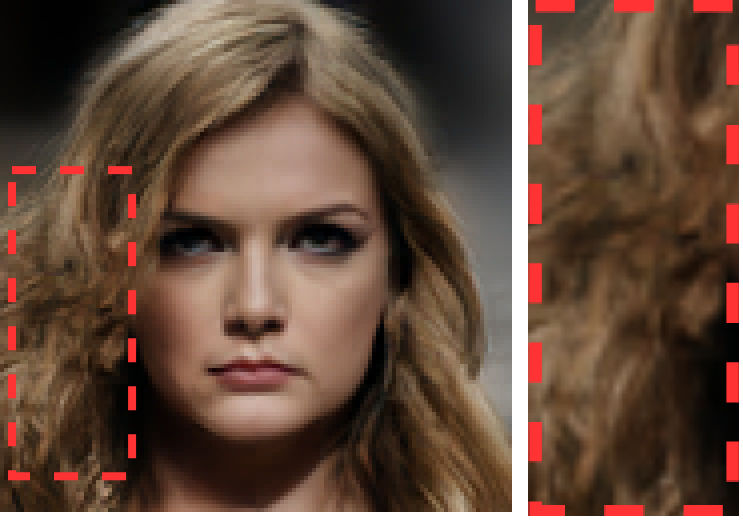}
      \subcaption{SR result w/o multi-scale frequency interrelations}
      \label{fig_mdwt_b}
    \end{minipage} \hspace{1mm}
    \begin{minipage}[t]{0.30\textwidth}
      \centering 
      \includegraphics[width=\textwidth]{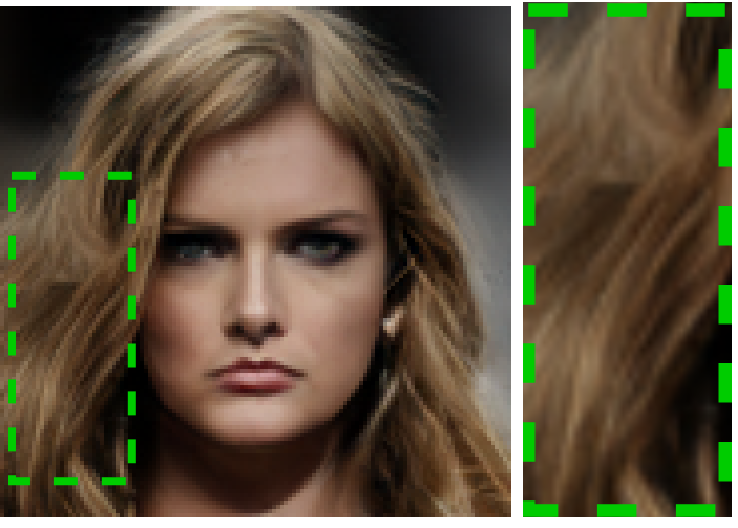}
      \subcaption{Ours (with multi-scale frequency interrelations considered)}
      \label{fig_mdwt_c}
    \end{minipage} \hspace{1mm}
    \begin{minipage}[t]{0.30\textwidth}
      \centering 
      \includegraphics[width=\textwidth]{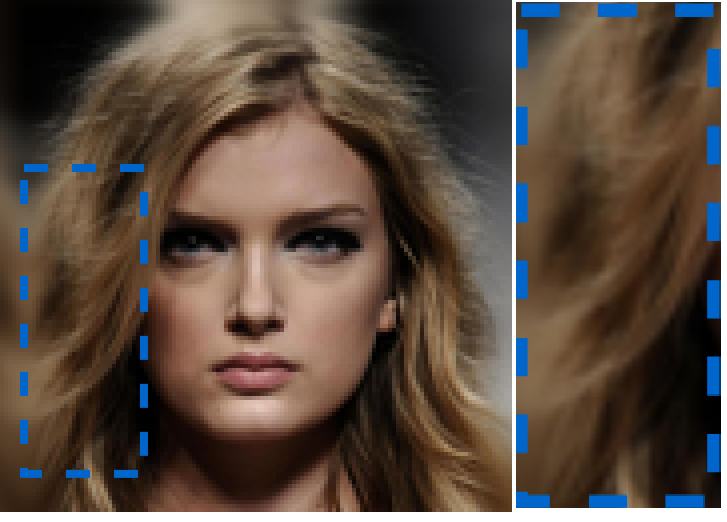}
      \subcaption{Ground truth}
      \label{fig_mdwt_d}
    \end{minipage}
  \end{minipage} 
  \vspace{-2mm}
  \caption{A transformer model based on multi-level wavelet spectra is explored for SR, enabling the learning of multi-scale frequency relationships to enhance SR results. Comparing (b) and (c), the proposed method produces more natural textures.}
  \vspace{-4mm}
  \label{fig_mdwt}
\end{figure}

Single-Image Super-Resolution (SISR) has gained growing attention for decades because of its broad application. It restores high-resolution (HR) images from the given low-resolution (LR) inputs, aiming at high performance on both objective fidelity and perceptual quality. Most methods establish the mapping from LR to HR images in image pixel domain. To capture the fine-grained frequency details critical for SR, some approaches use the Discrete Wavelet Transform (DWT) to convert images into the frequency domain. DWT depicts an image by a series of frequency sub-bands. The low-frequency (LF) sub-band reflects image global topology and affects objective fidelity, while the high-frequency (HF) sub-bands represent image textural details and affect perceptual quality significantly~\cite{WDST, WGSR}. 
As indicated in~\cite{WaveletSRNet}, SISR can be formulated as a wavelet coefficients prediction task. With correct predicting of wavelet coefficients based on LR input, the HR image can be reconstructed via inverse DWT (IDWT). Explicit optimization of wavelet coefficients in the frequency domain shows enhanced generation quality~\cite{WaveletSRNet, WGSR}. 

Plenty of works were proposed to improve the prediction accuracy of wavelet coefficients. Wavelet-SRNet~\cite{WaveletSRNet} adopts $N$ independent CNN subnets to predict multi-level of wavelet coefficients in parallel. WaveFace~\cite{WaveFace} employs a U-Net based model to recover the HF sub-bands sequentially during the upsampling process. WFEN~\cite{WFEN} takes DWT on each feature level in U-Net model to mitigate feature distortion during downsampling process.
Nevertheless, to the best of our knowledge, existing methods generally treat each level of HF coefficients independently, without considering the interrelations among the multi-scale HF sub-bands. 

Taking the objective of SISR into account, not only the spatial topology is important to avoid spatial distortion, the correlation among image frequency sub-bands is also crucial for better perceptual quality. Hence in this work, we explore a transformer model grounded on multi-level wavelet spectra for SISR, leveraging the excellence of transformer in modeling complex long-range relationships. Our model enables to uncover the correlations among the multiple scale frequency sub-bands, leading to elaborate textural details (as illustrated in Fig.~\ref{fig_mdwt} (b) \vs(c)).

Specially, we adopt Mallat decomposition~\cite{Mallat} for multi-level Discrete Wavelet Transform (MDWT), which decomposes the LF sub-band repeatedly at each subsequent level (detailed in Sec.~\ref{sec:preliminary}). The obtained frequency sub-bands are then combined together as a wavelet spectra representation of the image. As shown in Fig.~\ref{fig_mdwt}, it consists of one LF sub-band and multiple HF sub-bands at different scales, containing distinct levels of textural information. We then split the wavelet representation into patches for token embedding, not only on LF sub-band, but also on HF sub-bands. Unlike conventional methods that partition images spatially, our method divides images from the viewpoint of both spatial and frequency domains, facilitating the learning the frequency relationships among sub-bands. In addition, we propose a pyramid tokenization method given the sparsity of HF sub-bands. It reduces the token numbers largely and saves computation in transformer calculation without compromising model performance.


Inspired by the outstanding capability of Diffusion Model (DM) in generating fine image details, we formulate our method by using the conditional diffusion framework, and propose a \textbf{D}iffusion \textbf{T}ransformer model based on image \textbf{W}avelet spectra for SISR, abbreviated as \textbf{DTWSR}.
DMs reverse a diffusion process iteratively to achieve high-quality mapping from randomly sampled Gaussian noise to target images, avoiding the instability and mode-collapse present in previous generative models~\cite{DDPM, DDGAN, WaveDiff}. Due to the distinct variances in MDWT sub-bands, particularly between the smooth LF and sparse HFs, it is challenging to use a unified transformer model to denoise both LF and HF sub-bands simultaneously. Therefore, we design a dual-decoder transformer model, one for generating the high-energy elementary contents in LF (named as LEDec) and the other one for generating the sparse HF details (named as HDDec). It should be noted that the elementary contents from LEDec is not equal to LF sub-band. The LF sub-band still has HF components, though quite few. (A simple understanding is that we can continue the wavelet transform on LF to peel off the included HF components.) Thus, HDDec is designed to produce both the multi-level HF sub-bands and the HF components of LF sub-band. On one hand, our design is able to capture the interrelations among multi-scale HF sub-bands. On other hand, it promotes the realignment between LF and HF sub-bands, achieving SR with improved fidelity and perceptual quality.

The main contributions of this paper are as follows:
\begin{itemize}
  \item We propose a diffusion transformer model based on image wavelet spectra for SISR. It enables to explore the correlations among multi-scale frequency sub-bands.
  \item We design a pyramid tokenization method for embedding the multi-scale wavelet spectra. It reduces the token number largely for efficient calculation.
  \item A dual-decoder model is designed to prevent the entanglement of smooth and sparse frequency distributions for better fidelity and finer details.
  \item Extensive experiments are conducted on key benchmarks for face and general image SR tasks. Our method exhibits state of-the-art qualitative and quantitative results with improved image fidelity and perceptual quality.
\end{itemize}

\section{Related work}
\label{sec:related}

SISR has achieved great progress with the development of deep learning, including both model architecture and training framework. To improve the visual quality, various generative models are applied to train SISR model, including GAN~\cite{ESRGAN, FSRGAN, RealESRGAN, realsr, VQFR}, flow models~\cite{SRFlow,FlowIE,HCFlow} and Diffusion Models (DM)~\cite{SinSR, WaveFace, ECDP, WaveDM, DiWa}. Our work applies diffusion transformer based on wavelet spectra for SISR.

\noindent{\textbf {Diffusion based SISR.}} \ \ 
DM is rising as a powerful solution for high-quality image generation. SR3~\cite{SR3} adapts conditional DMs by concatenating upsampled LR with noisy HR image to perform SISR task. To speed up convergence and stabilize the training of DMs, SRDiff \cite{SRDiff} introduces residual prediction to speed up the convergence of DMs. ResDiff \cite{ResDiff}  uses a CNN network for initial recovery and then refines textural details by DM. IDM \cite{IDM} and ASIG \cite{ASIG} explore DMs in continuous SISR by integrating  implicit neural representation. ResShift \cite{Resshift} and SinSR \cite{SinSR} accelerate the inference speed of DM by modifying its sampling process and using knowledge distillation, respectively.


\noindent{\textbf {Discrete Wavelet Transform (DWT) based SISR.}} \ \ 
DWT has been used widely in SISR given its ability to express frequency information~\cite{LMP,WDST,FSR,DiWa,WDRN,WGSR}.  DWSR~\cite{DWSR} and DiWa~\cite{DiWa} are built on single-level DWT to improve model on precise textural details. Wavelet-SRNet~\cite{WaveletSRNet} and JWN~\cite{JWN} uses multi-branch CNN layers to predict wavelet coefficients based on LR input.  WaveMixSR~\cite{WaveMixSR}, WTRN~\cite{WTRN} and WFEN~\cite{WFEN} apply DWT on the extracted image features to complement HF information in SISR. WaveFace~\cite{WaveFace} and WaveDM~\cite{WaveDM} leverages the exponential shrinking of image size after DWT to reduce the computation burden of DM. Deng~\etal~\cite{WDST} proposed wavelet domain style transfer to achieve better perception-distortion (PD) trade-off for SISR. PDASR~\cite{PDASR} and WGSR~\cite{WGSR} optimize the loss on wavelet sub-bands to improve PD trade-off.


\noindent{\textbf {Transformer based SISR.}} \ \
Transformer based models are explored in SISR given its long-range modeling ability.
SwinIR~\cite{swinir} applies Swin transformer for image restoration. SwinFIR~\cite{zhang2022swinfir} improves SwinIR by incorporating Fourier Convolution to capture global information. HAT~\cite{HAT} combines self-attention, channel attention and overlapping cross-attention to active more pixel for better SR. Restormer~\cite{Restormer} proposes to perform self-attention in channel direction to capture long-range pixel interactions and achieves high performance in image restoration. 
LMLT~\cite{LMLT} divides image features along the channel dimension and employs attention with varying feature sizes to capture both local and global information. These works are based on pixel domain images. DWT is often employed in attention blocks of transformer model to enhance image feature, like~\cite{LIWT, WFEN, WTT, WAT}. To our knowledge, we are the first to model image's multi-scale wavelet spectra by using the basic transformer architecture for SR task.

\section{Discrete wavelet transform}
\label{sec:preliminary}
The discrete wavelet transform (DWT) is widely used to decompose an image into LF and HF sub-bands, especially the Haar wavelet \cite{Haar} used in this paper. 

Given a pixel image $I\in \mathbb{R} ^{H\times W\times 3}$, we decompose it by DWT operation ($\mathrm{DWT(\cdot)}$), and thus the low-frequency sub-band $x_L\in \mathbb{R} ^{\frac{H}{2} \times \frac{W}{2}\times 3}$ and high-frequency sub-bands $\{x_V,x_H,x_D\}\in\mathbb{R} ^{\frac{H}{2}\times \frac{W}{2}\times 3}$ can be produced:
\vspace{-2mm}
\begin{equation}
    \label{eq_dwt}
    x_L^1,x_V^1,x_H^1,x_D^1 = \mathrm{DWT}(I).
\end{equation}
The process can be conducted once more on $x_L^1$, resulting in
\vspace{-1mm}
\begin{equation}
    \label{eq_dwtj}
    x_L^2,x_V^2,x_H^2,x_D^2 = \mathrm{DWT}(x_L^1).
\end{equation}
By continuing the process, we have $\{x_L^J,x_V^J,x_H^J,x_D^J\} \in \mathbb{R}^{\frac{H}{2^J}\times \frac{W}{2^j}\times 3}$ after the $J$-th DWT. 

Replacing the LF sub-band recursively by the decomposed sub-bands in each level~\cite{Mallat}, the outputs after $J$-th DWT are $\{x_L^J,x_V^J,x_H^J,x_D^J,x_V^{J-1},...,x_D^1\}$. We reshape the multi-level sub-bands together and form a $J$-level wavelet spectrum representation of the image, denoted as $I_J^{fre}$, \ie, 
\vspace{-1mm}
\begin{equation}
    \label{eq_mdwtj}
    I_J^{fre} = \mathrm{MDWT}(I,J).
\end{equation}
Reversibly, the pixel image $I$ can be reconstructed via $J$-th invert DWT, (denoted as IMDWT):
\vspace{-2mm}
\begin{equation}
    \label{eq_idwtj}
    I = \mathrm{IMDWT}(I_J^{fre},J).
    \vspace{-2mm}
\end{equation}
An example is presented in Fig.~\ref{fig_mdwt_a}.

\begin{figure*}[!t]
  \centering
  \begin{minipage}[c]{0.9\textwidth}
    \centering
    \includegraphics[width=\textwidth]{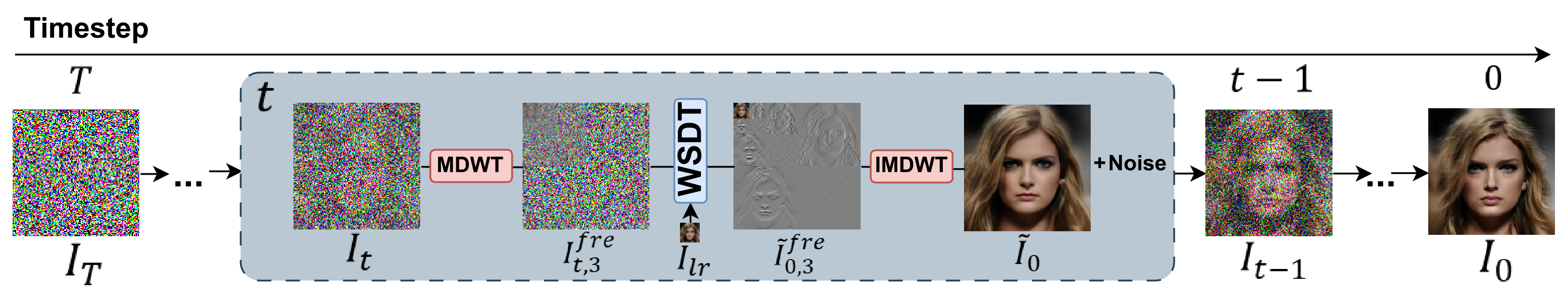}
    \subcaption{Overall conditional denoising process based on wavelet spectra for SISR. A 3-level multi-level MDWT is used as an example.}
    \label{fig_model_a}
  \end{minipage}\\
  \begin{minipage}[c]{0.99\textwidth}
    \centering 
    \includegraphics[width=\textwidth]{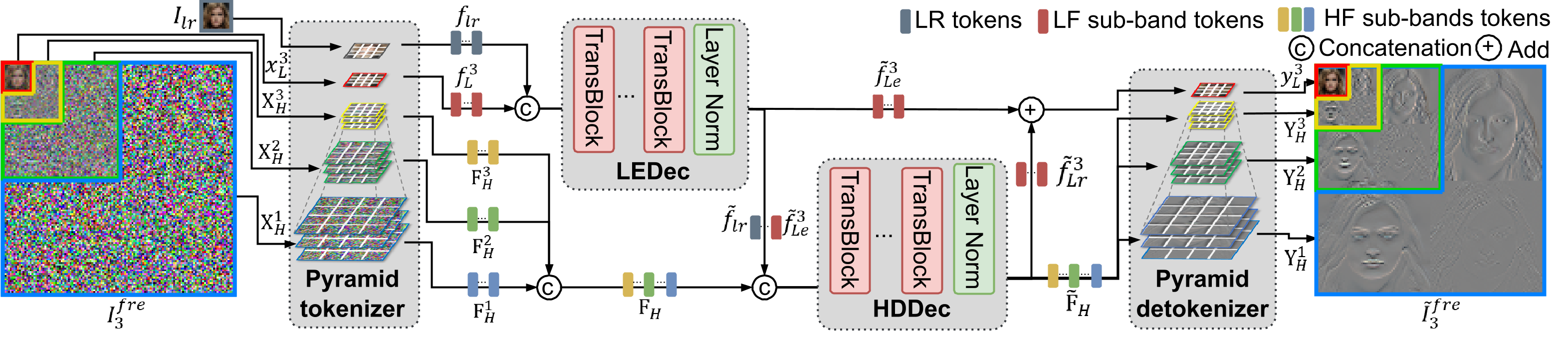}
    \subcaption{Detailed illustration of WSDT in (a), where the timestep condition and details of decoders are simplified for conciseness.}
    \label{fig_model_b}
  \end{minipage}
  \caption{Overview of the DTWSR framework. (a) shows SR sampling process, which follows the classic conditional denoising process of diffusion model (DM). The gray box shows how conditional DM on wavelet spectra is applied in each step. (b) illustrates the detailed structure of the proposed denoising network WSDT. The spectra image is embedded by a pyramid tokenizer. LEDec denoises the LF sub-band to obtain ${\tilde{f}}_{Le}^3$ under the guidance of the LR image. HDDec decodes the HF sub-bands tokens $\tilde{\mathbf{F}}_H$ and refines ${\tilde{f}}_{Le}^3$ by adding LR residual ${\tilde{f}}_{Lr}^3$, conditioning on LR features. Finally, the pyramid detokenizer transforms LF and HF tokens into the denoised spectrum ${\tilde{I}}_{3}^{fre}$.}
  \label{fig_model_overview}
\end{figure*}

\section{Methodology}
\label{sec:methodology}

In this section, we will introduce our Diffusion Transformer model based on Wavelet spectra for SR task.

\subsection{Conditional DM on wavelet spectra for SISR}
DM is a parameterized Markov chain that produces samples matching the training data distribution. It consists of a forward diffusion process and a reverse denoising process. The diffusion process gradually adds Gaussian noise to a clean image according to a pre-defined Markov process, while the denoising process recovers the clean image from Gaussian noise by removing noise iteratively via a denoising network learned from the diffusion process. For SR task, it is request that the recovered image is consistent with the content from the given LR input, resulting in a conditional DM:
\vspace{-1mm}
\begin{equation}
  \label{CDM}
  p_\theta({I}_{t-1}|{I}_t,{I}_{lr}) = \mathcal{N}({I}_{t-1}; \mathbf{\mu}_{\theta}({I}_t,t,{I}_{lr}), \mathbf{\Sigma}_\theta({I}_t,t,{I}_{lr})),
\end{equation}
where ${I}_{lr}$ denotes the LR input, $\theta$ is the parameter of our designed Wavelet Spectrum Denoising network with Transformer, named as \textbf{WSDT}, and $t$ is the denoising step. By refining ${I}_{t}$ recursively conditioning on ${I}_{lr}$, the SR image $I_{0}$ can be obtained. The process is illustrated in Fig.~\ref{fig_model_a}. 

Different from previous methods that remove noise from $I_{t}$ in pixel domain, we attempt to leverage the wavelet frequency spectrum to improve the generation quality. Hence we transform the pixel image $I_{t}$ to a $J$-level wavelet spectrum representation $I_{t,J}^{fre}$ for refinement (as shown in Fig.~\ref{fig_model_a}). The included LF sub-band and the set of HF sub-bands are denoted as $x_{t,L}^J$ and $\mathbf{X}_{t,H} = \{\mathbf{X}_{t,H}^j\}$ separately, where $\mathbf{X}_{t,H}^j=\{x_{t,V}^j,x_{t,H}^j,x_{t,D}^j\}, j\in \{1,...,J\}$.


In our model, we set the level of MDWT according to the 
magnification of SR. For a upscaling factor $N$, we perform $J$-level DWT, with $J=\mathrm{ceil}(\mathrm{log}_2 N)$. With this setting, the size of LF sub-band will be no larger than that of LR input. It would be relatively easier to learn the mapping between two similar size images with similar distribution.


\subsection{Wavelet Spectrum Denoising Network with Transformer (WSDT)}
Fig.~\ref{fig_model_b} presents the architecture of WSDT. Given the noisy image in wavelet spectrum $I_{t,J}^{fre}$, we firstly patchify it into a sequence of tokens. Then dual transformer decoders are designed to denoise the elementary contents in LF and the multi-scale HF details respectively, with in-context conditioning on LR input. The transformer blocks will learn the interrelations both in spatial domain and among multi-scale frequencies, leading to more accurate denoising across sub-bands. 
In this section, we omit the time step $t$ for simplicity as the operations are the same for each time step.

\subsubsection{Pyramid tokenization}
\label{subsection:pyramid}
\begin{figure}
  \centering
  \begin{minipage}[c]{0.26\textwidth}
    \centering
    \includegraphics[width=\textwidth]{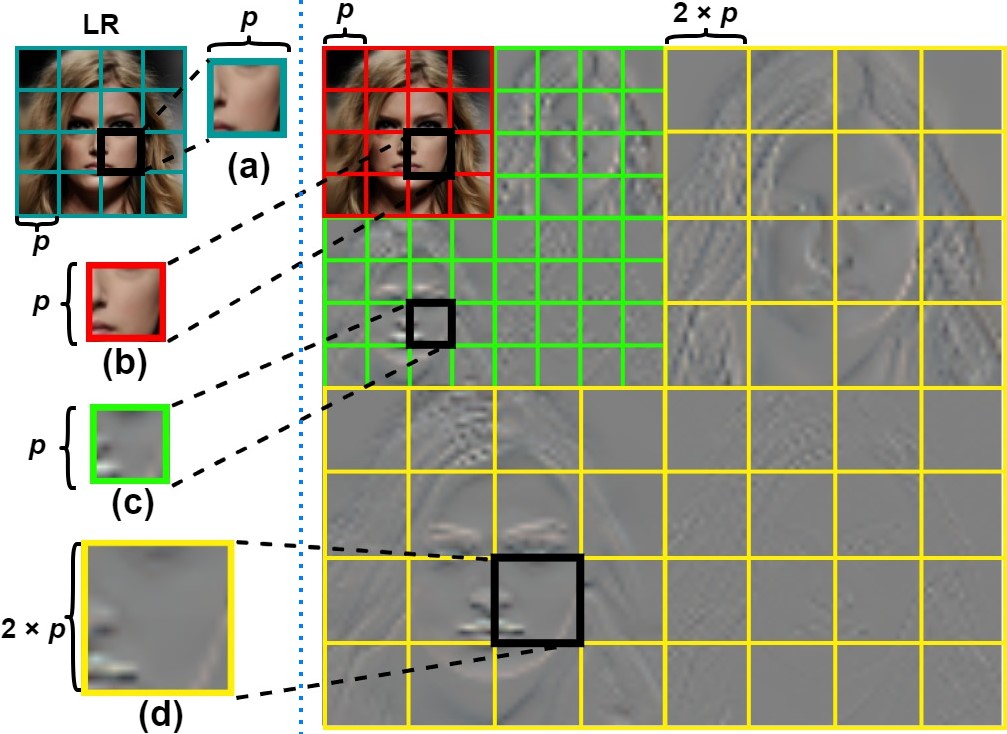}
    \subcaption{Pyramid patch sizes}
    \label{fig_pyramidtoken_a}
  \end{minipage}
  \hspace{1mm}
  \begin{minipage}[c]{0.18\textwidth}
    \centering
    \includegraphics[width=\textwidth]{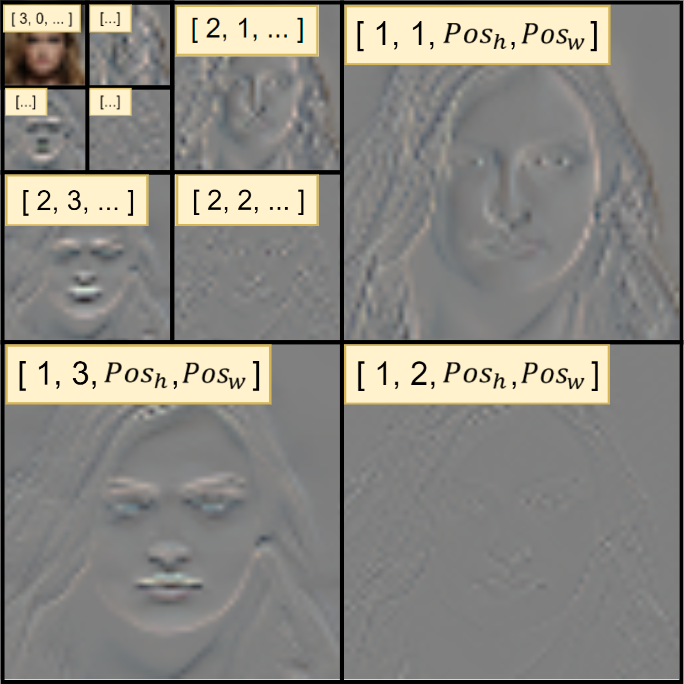}
    \subcaption{4D positional encoding}
    \label{fig_pyramidtoken_b}
  \end{minipage}
  \vspace{-2mm}
  \caption{Illustration of pyramid tokenization. Our method enables consistent receptive fields across frequency sub-bands.}
  \vspace{-4mm}
  \label{fig_pyramidtoken}
\end{figure}

Conventional methods split image into same size patches for embedding~\cite{VIT}. Considering the sparsity of HF 
components in wavelet spectrum image $I_{J}^{fre}$, we design a pyramid tokenization method. The LF sub-band is divided using a smaller patch size, while the sparse HF sub-bands are split by a larger patch size, as shown in Fig. \ref{fig_pyramidtoken_a}.
Moreover, in order to keep consistent receptive fields across different level of sub-bands, we define the pyramid patch size $p^j$ according to its levels $j$ in MDWT:
\vspace{-2mm}
\begin{equation}
\label{eq_psize}
\vspace{-1mm}
p^j =p_{min} \times 2^ {J-j} , j\in \left\{1,...,J\right\},
\end{equation}
where $p_{min}$ is the patch size for LF sub-band.

The pyramid tokenization is achieved by convolutional layers $\mathrm{Conv2d}(\cdot)$, with the kernel size and stride set to be $p^j$. For each level of $\mathbf{X}_{H}^j$, we concatenate them together for embedding. LR input and LF sub-band are embedded using separate CNN layers. The resulted image tokens are denoted as $\{{f}_{lr},{f}_L^J,\mathbf{F}_H\}$, $\mathbf{F}_H = \{\mathbf{F}_H^j\}$, $j\in\{J,...,1\}$:
\vspace{-2mm}
\begin{equation}
  \label{eq_pembed}
  \begin{aligned}
    &{f}_{lr}=\mathrm{Conv2d}_{lr}(I_{lr}), {f}_L^J =\mathrm{Conv2d}_L(x_L^J), \\
    &\mathbf{F}_H^j=\mathrm{Conv2d}^j_H(\mathbf{X}_H^j),j \in\left\{J,...,1\right\}.\\
  \end{aligned}
\end{equation}

Next, we define the position embedding for each token. With our pyramid tokenization, the resulted tokens have the same 2D-absolute position $[Pos_h,Pos_w]$ in each sub-band, which makes it easier to learn the relations among sub-bands. To distinguish the level of wavelet spectrum and the specific sub-band in each level, we additionally specify the level $j\in \left\{1,...,J\right\}$ and the sub-band position $d \in \{x_L=0,x_V=1,x_D=2,x_H=3\}$, leading to a 4D position $[j,d,Pos_h,Pos_w]$ for each token, as the example shown in Fig. \ref{fig_pyramidtoken_b}. The 4D position is encoded by standard ViT frequency-based positional embeddings (the sine-cosine version) \cite{VIT} and then added to the patch embeddings to retain positional information.

\begin{figure}
  \centering
  \begin{minipage}[c]{0.20\textwidth}
    \centering
    \includegraphics[width=\textwidth]{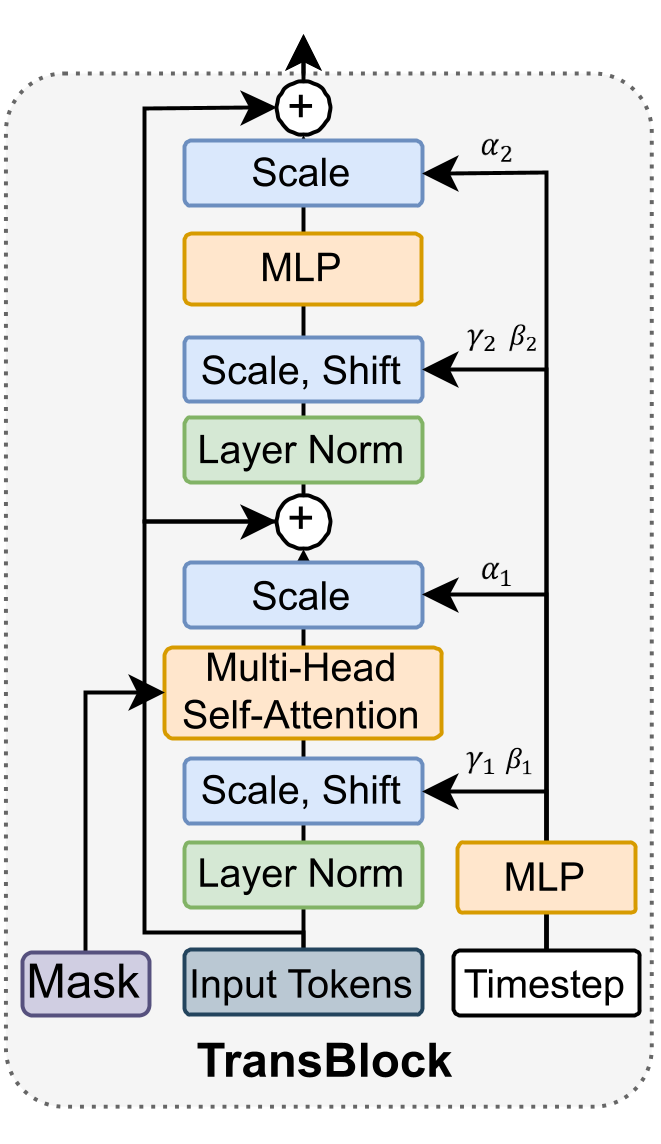}
    \subcaption{TransBlock architecture}
    \label{fig_block}
  \end{minipage} 
  \hspace{5mm}
  \begin{minipage}[c]{0.16\textwidth}
    \centering
    \begin{minipage}[c]{1\textwidth}
      \centering
      \includegraphics[width=\textwidth]{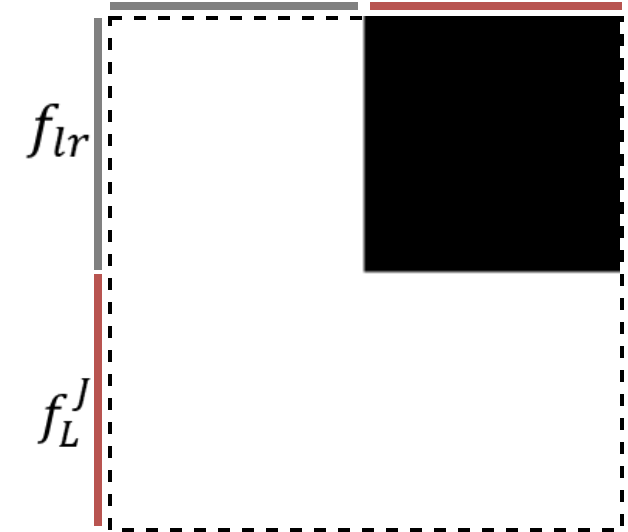}
      \subcaption{Self-attention mask $M_{low}$ in LEDec}
      \label{fig_mask_a}
    \end{minipage} \\\vspace{1mm}
    \begin{minipage}[c]{1\textwidth}
      \centering 
      \includegraphics[width=\textwidth]{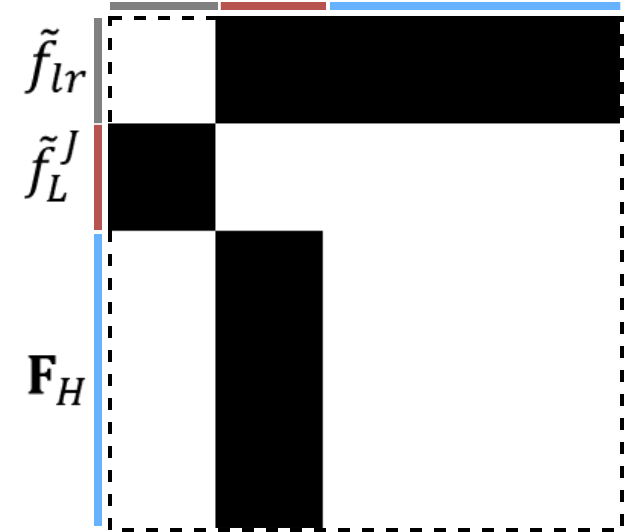}
      \subcaption{Self-attention mask $M_{high}$ in HDDec}
      \label{fig_mask_b}
    \end{minipage}
  \end{minipage} 
  
  \caption{Illustration of TransBlock and the designed masks in decoders. The black parts correspond to the masked tokens in self-attention computation. \textcolor[rgb]{0.5,0.5,0.5}{Gray} lines indicate LR tokens. \textcolor[rgb]{0.72,0.33,0.31}{Red} lines indicate LF sub-band tokens. \textcolor[rgb]{0.4,0.69,1}{Blue} lines indicate HF sub-bands tokens.}
  \vspace{-4mm}  
  \label{fig_masks}
\end{figure}

\subsubsection{Dual-decoder design}
\label{subsection:condition}
Given that $x_{L}^J$ and $\{\mathbf{X}_{H}^j\}$ have different distributions, using a unified decoder to denoise spectra with variant distributions is difficult. Thus, we design dual transformer decoders. 
Rather than denoising $x_{L}^J$ and $\{\mathbf{X}_{H}^j\}$ separately as in previous methods~\cite{WaveFace, WaveDM}, we handle their components more carefully.
Considering the HF component left in $x_{L}^J$, we use one decoder to denoise the smooth elementary contents in $x_{L}^J$. The decoder is named as LF Elementary Decoder (\textbf{LEDec}). The other one will denoise all the HF coefficients $\{\mathbf{X}_{H}^j\}$ as well as the left HF components in $x_{L}^J$ (denoted as LF Residual). The decoder is named as HF Detail Decoder (\textbf{HDDec}).
Both decoders are composed of multiple transformer blocks~\cite{DiT} as denoted by TransBlock in Fig.~\ref{fig_model_overview}. The detailed structure of each TransBlock is presented in Fig.~\ref{fig_block}. 

\noindent{\textbf {LEDec.}} \ \ 
As shown in Fig.~\ref{fig_model_b}, LEDec aims to denoise the smooth elementary component in $x_{L}^J$. We use the in-context conditioning method~\cite{DiT} to incorporate the information from LR input. In particular, we concatenate the LR tokens $f_{lr}$ and noised LF tokens $f_L^J$ as input to LEDec. To prevent LR condition from being contaminated by the noised LF tokens, we tailor the attention mask $M_{low}$ in self-attention computation shown in Fig.~\ref{fig_mask_a}. The time step $t$ is also embedded and participates in the denoising process via Adaptive layer norm zero (AdaLN-Zero) manner~\cite{DiT} in each TransBlock. The process can be formulated as:
\vspace{-1mm}
\begin{equation}
  \label{eq_lfdec}
  \tilde{f}_{lr},\tilde{f}_{Le}^J=\mathrm{LEDec}([f_{lr},f_L^J],M_{low},t),
\end{equation}
where $[\cdot,\cdot]$ is the concatenation operation. 


\noindent{\textbf {HDDec.}} \ \ 
As depicted in Fig.~\ref{fig_model_b}, the inputs to HDDec include the encoded LR and LF tokens, as well as the noised HF tokens $\mathbf{F}_H$. HDDec denoises the multi-level HF sub-bands $\{\mathbf{X}_{H}^j\}$ and LF Residual. The denoising of LF Residual not only supplements the HF components in $x_{L}^J$, but also promotes realignment of LF and HF sub-bands, contributing to finer image generation. We use in-context conditioning manner as well for embedding LR information. A tailored attention mask $M_{high}$ is designed as illustrated in Fig.~\ref{fig_mask_b}, which will avoid the unnecessary interaction among tokens. The operations are formulated as 
\begin{equation}
  \label{eq_hfdec}
  \begin{aligned}
    \hat{f}_{lr}&,{\tilde{f}_{Lr}^J},\mathbf{\tilde{F}}_H=\mathrm{HDDec}([\tilde{f}_{lr},\tilde{f}_{Le}^J,\mathbf{F}_H],M_{high},t),
  \end{aligned}
\end{equation}
where $M_{high}$ is worth to be noticed: (1) LR tokens are invisible to LF tokens to avoid LR conditioning in HDDec to be used for LF sub-band denoising and force the re-alignment between LF and HF sub-bands. (2) LF tokens are invisible to HF tokens to prevent the influence from LF tokens on HF. 

\noindent{\textbf{Detokenization.}} \ \ 
Finally, the obtained tokens will be detokenized into wavelet coefficients. For LF sub-band,  $\tilde{f}_{Le}^J$ and $\tilde{f}_{Lr}^J$ are added together for detokenizing, while HF coefficients are detokenized from $\mathbf{\tilde{F}}_H = \{\mathbf{\tilde{F}}_H^j\}$ respectively. We apply a layer norm (AdaLN) to incorporate the time step $t$, and decode each token linearly by $\mathrm{FC}(\cdot)$ into a $p^j\times p^j \times c$ tensor, $c$ is the channel number of that spectrum \ie,  
\vspace{-2mm}
\begin{equation}
  \label{eq_pembedout}
  \begin{aligned}
    &y_L^J=\mathrm{FC}_L(\tilde{f}_{Le}^J+\tilde{f}_{Lr}^J,t) 
    \\
    &\mathbf{Y}_H^j=\mathrm{FC}^j_H(\mathbf{F}_H^j,t),j \in\left\{1,...,J\right\},
    \vspace{-2mm}
  \end{aligned}
\end{equation}

The output tensors are then rearranged according to their original spatial layout, resulting $\tilde{I}_J^{fre} = \{y_L^J, \mathbf{Y}_H \}$, $\mathbf{Y}_H=\{ \mathbf{Y}_H^j\}, j \in\left\{1,...,J\right\}$. It is transformed to pixel image by inverse wavelet transform, \ie, $\tilde{I} = \mathrm{IMDWT}(\tilde{I}_J^{fre}, J)$.

\subsubsection{Optimization.} 
To accelerate the denoising process, we adopt the optimization method proposed in DDGAN~\cite{DDGAN}. It introduces a time-dependent discriminator to learn data distribution at a large denoising step (which is no longer Gaussian), enabling fast sampling without affecting model convergence. Specifically, let $I_0$ be the clean HR image and $I_t$ be a noised image at timestep $t$ sampled from the diffusion process $q(I_t|I_0)$:
\vspace{-1mm}
 \begin{equation}
\label{diffusion}
q(I_t|I_0) = \mathcal{N}(I_t; \sqrt{\alpha_t} I_0, (1-\alpha_t) \mathbf{I}),
\vspace{-1mm}
\end{equation} 
where $\alpha_t$ is predefined according to noise schedule.
 
During the denoising process, our network outputs an denoised image $\tilde{I}_0$ in each step, which is an approximation of $I_0$. A perturbed sample $\tilde{I}_{t-1}$ can be derived by Eq.~(\ref{diffusion}). DDGAN trains a discriminator $D(\cdot)$ to 
distinguish the real pairs $(I_{t-1},I_t)$ and the fake pairs $(\tilde{I}_{t-1},I_t)$ adversarially, formulated as:
\vspace{-1mm}
\begin{equation}
  \label{eq_trainD}
  \begin{aligned}
   & \mathcal{L}_{adv}^D = -\mathrm{log}({D}(I_{t-1},I_t,t))+\mathrm{log}({D}(\tilde{I}_{t-1},I_t,t)). \\
   & \mathcal{L}_{adv}^G = -\mathrm{log}(\mathrm{D}(\tilde{I}_{t-1},I_t,t)).
  \end{aligned}
  \vspace{-1mm}
\end{equation}

To preserve the consistency of wavelet sub-bands without losing of frequency details, we build reconstruction term by $L_1$ loss in both pixel and frequency domain:
\vspace{-1mm}
\begin{equation}
  \label{eq_trainR}
  \begin{aligned}
   & \mathcal{L}_{pixel} = ||\tilde{I}_{0}-I_0||,
   & \mathcal{L}_{fre} =||\tilde{I}_{0,J}^{fre}-I_{0,J}^{fre}||.
  \end{aligned}
\end{equation}
The overall objective of the generator is 
\vspace{-1mm}
\begin{equation}
  \label{eq_trainAll}
  \begin{aligned}
    \mathcal{L}^G = \alpha \mathcal{L}_{adv}^G + \beta \mathcal{L}_{pixel} + \gamma \mathcal{L}_{fre},
  \end{aligned}
  \vspace{-1mm}
\end{equation}
where $\alpha, \beta, \gamma $ are adjustable weighting hyper-parameters.



\section{Experiments}
\label{sec:experiments}

\subsection{Implementation details}

\noindent{\textbf {Datasets.}} \ \ 
We evaluate our method on face and general scene datasets.
For face SISR, we train the proposed DTWSR on FFHQ~\cite{FFHQ} and evaluate on CelebA~\cite{Celeba} validation set. For general scene SISR, we use DIV2K~\cite{DIV2K} and Flicker2K~\cite{Flick2K} for training and test the model on several datasets including DIV2K validation set, Manga109~\cite{Mang}, Set5~\cite{Set5} and Set14~\cite{Set14}. Moreover, we evaluate our method for real-world image restoration task~\cite{Resshift} to show its generalization capability, where the model is trained on ImageNet~\cite{ImageNet} and test on RealSR dataset~\cite{realsr}.

\noindent{\textbf {Training details.}} \ \ 
Our implementation is mainly based on DDGAN~\cite{DDGAN} and DiT~\cite{DiT}. We adopt the same training configurations as DDGAN and DiT for all experiments. 
Training epochs are set as 250 (about 0.5M iterations with a batch size of 32) for face SISR and 300 (about 0.8M iterations with a batch size of 32) for general SISR (see Supplement for more hyperparameters and relevant details).

\noindent{\textbf {Evaluation metrics.}} \ \ 
For evaluation, we adopt two distortion-based metrics PSNR and SSIM~\cite{SSIM}, as well as perception-based metrics FID~\cite{FID} and LPIPS~\cite{LPI}. Additionally, we adopt the identity similarity (denoted as "Deg.")~\cite{VQFR} and consistency scores~\cite{SR3} (denoted as "Cons.") to measure the fidelity of outputs. Deg. means the face identity distance with angles between the restored image and ground-truth extracted by ArcFace~\cite{ArcFace}. Cons. measures the mean squared error (MSE) ($\times10^{-5}$) between the down-sampled outputs and the LR image~\cite{SR3}. 

\subsection{Comparison with state-of-the-art methods}

\begin{figure}[h]
\centering
\vspace{-1mm}
\includegraphics[width=0.95\columnwidth]{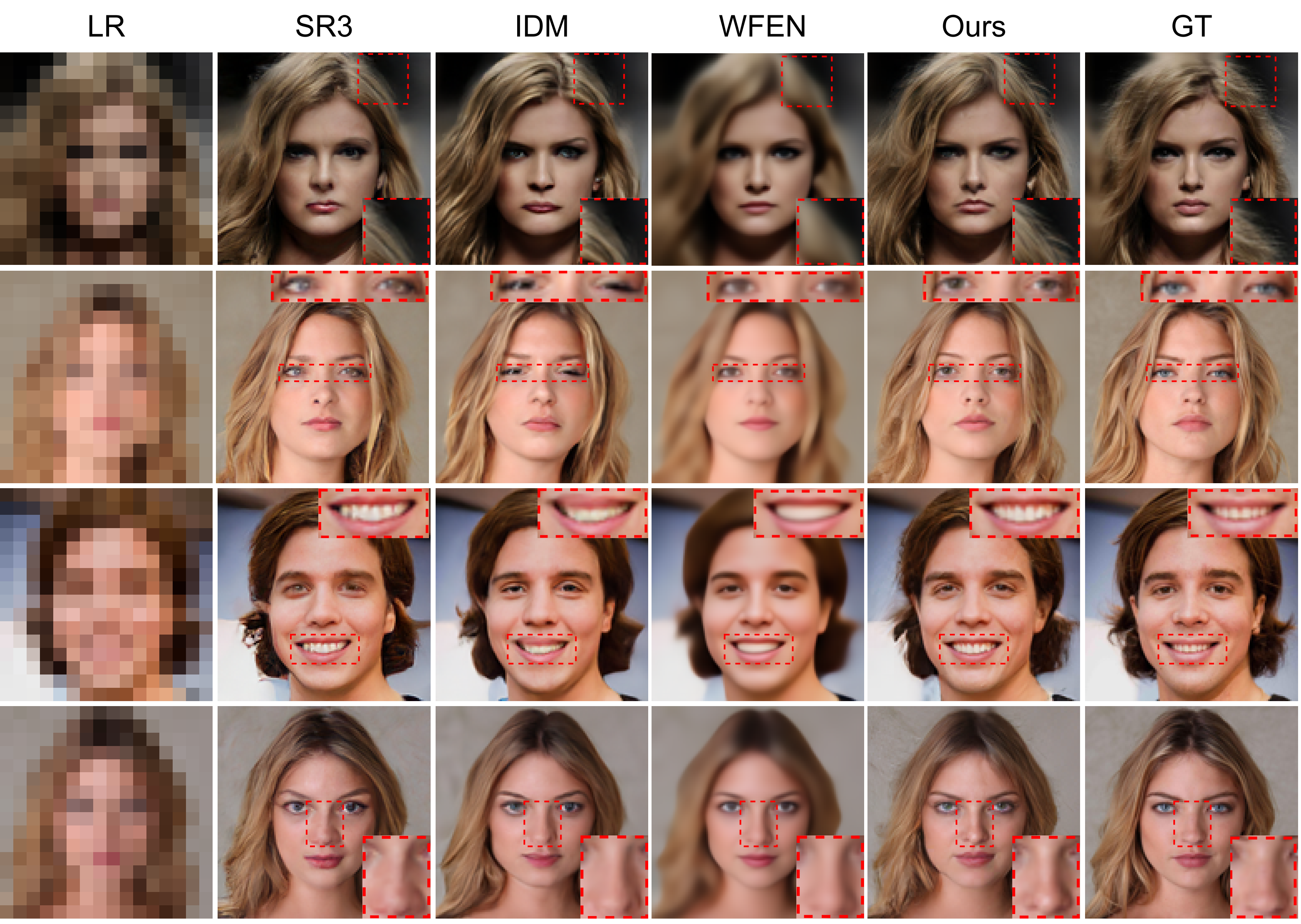}
\vspace{-1mm}
\caption{Qualitative comparison on $8\times$ SISR on CelebA-HQ. Our results not only maintain higher fidelity and more credible identities (eyes, mouth, \etc.) close to the ground-truth, but also have finer textual details (skin, hair, \etc.). Zoom in for best view.}
\label{fig_facex8}
\vspace{-5mm} 
\end{figure}

\noindent{\textbf {Face SISR.}} \ \ 
Following IDM~\cite{IDM}, we evaluate the proposed DTWSR on 100 face images from CelebA-HQ. Images are super-resolved from $16^2$ to $128^2$ pixels with $8\times$ upscaling. As illustrated in Table \ref{tab_facex8}, our method shows promising quantitative results compared with SOTA. It achieves the lowest Cons. and Deg., indicating that DTWSR can preserve more face attributes to maintain the authenticity of outputs. The lowest FID score demonstrates its finer and more realistic detail generation. WFEN~\cite{WFEN}, trained with MSE loss merely, estimates the posterior mean effectively without considering data distribution, yielding over-smoothed images with poor perceptual quality. As visualized in Fig.~\ref{fig_facex8}, our method provides much more natural and realistic results with rich details such as exquisite skin, teeth and hair texture. In addition, our results are more similar to the ground-truth (\ie, eyes, mouth and nose), without spatial distortion, leading to better objective fidelity.


\begin{table}[t]
  \caption{Quantitative comparison with several baselines on $16^2$ to $128^2$ face SISR. The best and second best results are highlighted in \textbf{bold} and \underline{underline}.}
  \centering
  \vspace{-1mm}
  \setlength\tabcolsep{3pt}
  \begin{tabular}{lccccc}
    \hline
    Method & PSNR$\uparrow$&SSIM$\uparrow$&Cons.$\downarrow$& Deg.$\downarrow$&FID$\downarrow$   \\ \hline
    FSRGAN~\cite{FSRGAN} & 23.01 & 0.62 & 33.8  & -     & -     \\
    SR3~\cite{SR3}   & 23.04 & 0.65 & 2.68  & 58.99 & 70.82 \\
    DiWa~\cite{DiWa}  & 23.34 & 0.67 & -     & -     & -     \\
    IDM ~\cite{IDM}  & 24.01 & \underline{0.71} & 2.14 & 58.07 & \underline{57.07} \\
    WFEN~\cite{WFEN}  & \textbf{25.53} & \textbf{0.77} & \underline{2.13} & \underline{57.96} & 106.34 \\ \hline
    Ours  & \underline{24.09} & \underline{0.71} & \textbf{0.50} & \textbf{53.85} & \textbf{56.77} \\ \hline
  \end{tabular}
  \label{tab_facex8}
\end{table}

\begin{figure*}
  \centering
  \includegraphics[width=2.0\columnwidth]{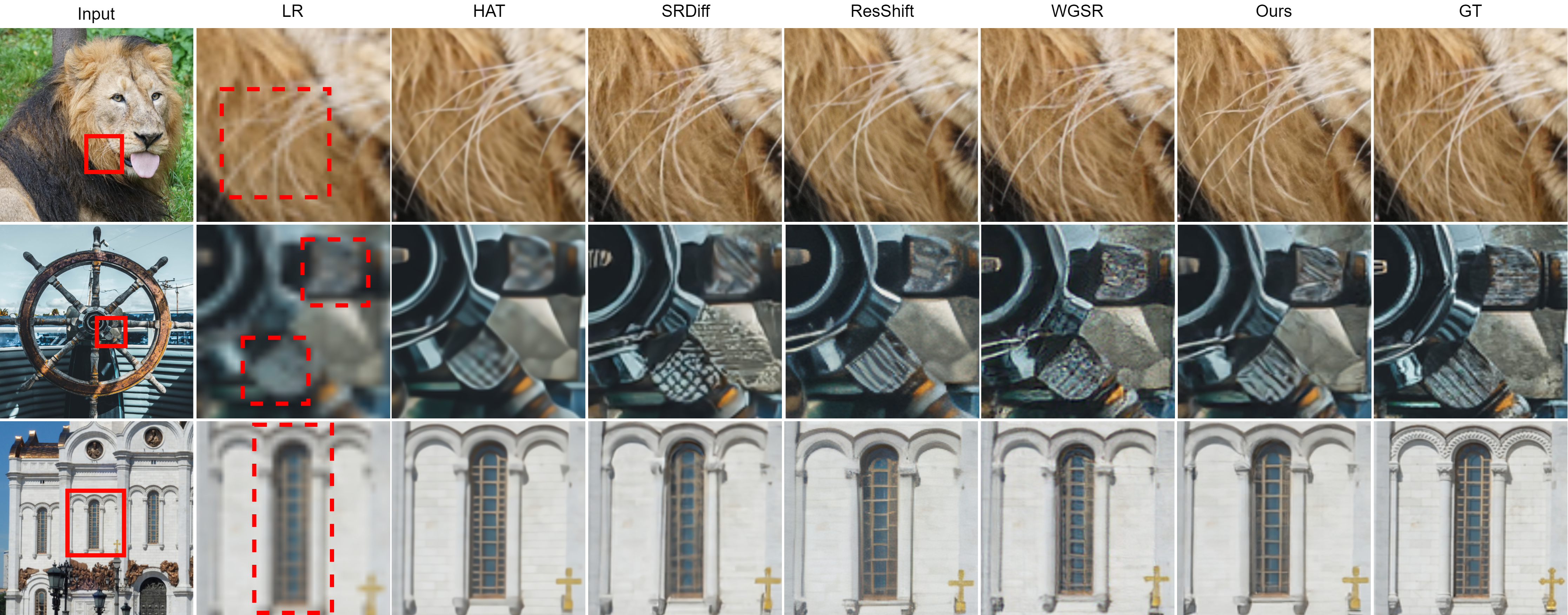}
  \vspace{-1mm}
  \caption{Qualitative comparison on $4\times$ SISR on DIV2K~\cite{DIV2K}. The parts for detailed comparison are marked with red boxes in the images. Our results provides more credible details than other methods. Zoom in for best view.}
  \vspace{-2mm}
  \label{fig_generalx4}
\end{figure*}

\noindent{\textbf {General scene SISR.}} \ \ 
Following common practice~\cite{IDM, Resshift, WGSR} for fair comparison, We perform $4\times$ SR on general scene SISR datasets. In Table \ref{tab_generalx4}, we evaluate DTWSR on DIV2K validation set and compare the results with various prior arts, including regression-based and generative methods. Regression-based approaches (LIIF~\cite{LIIF} and HAT~\cite{HAT}) yield higher PNSR and SSIM scores but worse LPIPS. our DTWSR shows better perception-distortion tradeoff~\cite{WDST} performance compared with other generative methods, with both higher reconstruction accuracy and better perceptual quality. The qualitative comparisons are in Fig.~\ref{fig_generalx4}. Our method produces correct object structure with rich textural details (see the clearer animal fur). The regression-based HAT suffers from the typical over-smoothing issue. SRDiff~\cite{SRDiff}, ResShift~\cite{Resshift} and WGSR~\cite{WGSR} are negatively affected by mis-alignment with LR condition, resulting in various artifacts (see the second and third rows). 

We conducted more comparison on Manga109~\cite{Mang}, Set5~\cite{Set5} and Set14~\cite{Set14} as shown in Table \ref{tab_generalother}. The proposed DTWSR outperforms generative methods on most metrics, which proves the effectiveness of our method further. 

\begin{table}[]
  \caption{Quantitative comparison with several baselines on $4\times$ general SISR. The best and second best results are highlighted in \textbf{bold} and \underline{underline} among generative models.}
  \vspace{-1mm}
  \begin{tabular}{lcccc}
  \hline
  Method  & PSNR$\uparrow$  & SSIM$\uparrow$ & Cons.$\downarrow$  & LPIPS$\downarrow$  \\ \hline
  Bicubic & 26.70  & 0.77 & 17.86 & 0.409 \\
  LIIF~\cite{LIIF}  & 29.29  &0.82  &0.820 & 0.132 \\
  HAT~\cite{HAT}  & 29.83  &0.87  &0.847 & 0.125 \\ \hline
  ESRGAN~\cite{ESRGAN} & 26.22 & 0.75 & 7.221 & 0.124 \\
  SRFlow~\cite{SRFlow} & 27.09 & 0.76 & -      & 0.120 \\ 
  SRDiff~\cite{SRDiff} & 27.41 & \textbf{0.79} & \underline{1.254}  & 0.136  \\
  IDM~\cite{IDM}    & 27.59 & \underline{0.78} & -      & -     \\
  DiWa~\cite{DiWa}   & \underline{28.09} & \underline{0.78} & -      & 0.104 \\
  ResDiff~\cite{ResDiff}& 27.94 & 0.72 & -      & -     \\
  ResShift~\cite{Resshift}     & 27.24 & 0.74 & 11.73 & 0.105 \\
  WGSR~\cite{WGSR} & 27.37  & 0.76  & 2.187  & \textbf{0.096}  \\
   \hline
  Ours    & \textbf{28.18} & \textbf{0.79} & \textbf{1.151}  & \underline{0.097} \\ \hline
  \end{tabular}
  \label{tab_generalx4}
\end{table}

\begin{table}[]
  \vspace{-2mm}
  \caption{Quantitative comparison on Manga109~\cite{Mang}, Set5~\cite{Set5} and Set14~\cite{Set14} dataset. The best and second best results are highlighted in \textbf{bold} and \underline{underline} among generative models.}
  \centering
  \vspace{-1mm}
  \setlength\tabcolsep{0.5pt}
  \begin{tabular}{lcccccc}
  \hline 
          & \multicolumn{2}{c}{Manga109 4$\times$} & \multicolumn{2}{c}{Set14 4$\times$} & \multicolumn{2}{c}{Set5 4$\times$} \\ \cline{2-7} 
          & PSNR$\uparrow$    & SSIM$\uparrow$    & PSNR$\uparrow$ & SSIM$\uparrow$  & PSNR$\uparrow$  & SSIM$\uparrow$    \\ \hline
  SRDiff~\cite{SRDiff}  & 27.04              & 0.813             & 25.63            & 0.702            & 28.72            & 0.843           \\
  SR3~\cite{SR3}    & 26.88              & 0.805             & 25.29            & 0.684            & 27.31            & 0.767           \\
  ResDiff~\cite{ResDiff} & \underline{27.76}              & \underline{0.832}            & \underline{26.19}          & \underline{0.718}            & \underline{29.32}            & \textbf{0.854}           \\
  ResShift~\cite{Resshift} & 26.91              & 0.824             & 25.11            & 0.682            & 28.54            & 0.817           \\
  WGSR~\cite{WGSR} & 26.59              & 0.823             & 25.28            & 0.644            & 27.65            & 0.781           \\\hline
  Ours    & \textbf{27.79}              & \textbf{0.865}             & \textbf{26.58}            & \textbf{0.725}            & \textbf{29.47}            & \underline{0.846}          \\ \hline
  \end{tabular}
  \label{tab_generalother}
  \vspace{-3mm}  
\end{table}

\begin{figure}
  \centering
  \includegraphics[width=1\columnwidth]{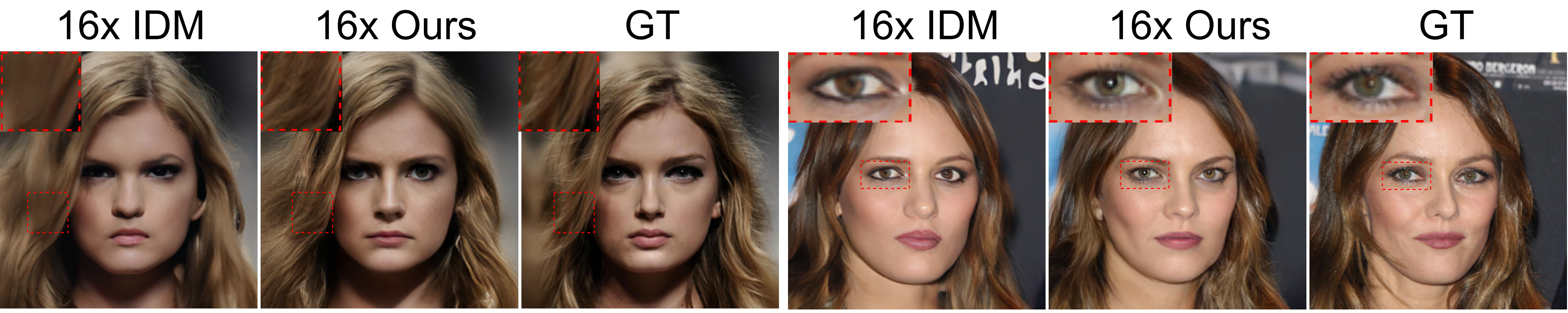}
  \vspace{-6mm}
  \caption{Qualitative comparison on $16\times$ SISR ($16^2$ to $256^2$) on CelebA-HQ~\cite{Celeba}. Our results not only maintain higher fidelity but also have finer textual details. Zoom in for best view. }
  \label{fig_compare_largescale}
\end{figure}

\begin{table}[]
  \vspace{-2mm}
  \caption{Quantitative comparison with SOTA method~\cite{IDM} on large-magnification face SISR. Our results obtain better scores.}
  \centering
  \vspace{-2mm}
  \setlength\tabcolsep{5pt}
  \begin{tabular}{lccccc}
    \hline
    Method & PSNR$\uparrow$&SSIM$\uparrow$&Cons.$\downarrow$& Deg.$\downarrow$&FID$\downarrow$   \\ \hline
    IDM-$12\times $ & 23.21 & 0.65 &  8.16 &  55.44 & 60.58 \\
    \textbf{Ours-}$12\times$ & 23.36 & 0.67 & 0.64  & 52.38 & 58.06 \\ \hline
    IDM-$16\times$   &23.15 & 0.65 &9.57  & 56.53 & 63.99 \\
    \textbf{Ours-}$16\times$ &23.18 & 0.65 & 0.89  & 53.54 & 60.37 \\ \hline
  \end{tabular}
  \label{tab_face_largescale}
\end{table}

\noindent{\textbf {Large-magnification SISR.}} \ \ 
Here we explore DTWSR on large-magnification SISR. We test DTWSR on $12\times$ ($16^2$ to $196^2$) and $16\times$ ($16^2$ to $256^2$) face SISR and compare the results with SOTA method IDM~\cite{IDM}\footnote{We train IDM on $12\times$ and $16\times$ face SISR based the official code.\label{fn:IDM}} 
As shown in Fig.~\ref{fig_compare_largescale} and Table \ref{tab_face_largescale}, IDM shows a significant drop in Cons. and FID under large-magnification SISR because of the less control on frequency layers. Our DTWSR maintain good performance on both fidelity and perceptual quality.

\begin{table}[]
  \vspace{-2mm}
  \caption{Quantitative comparison with latest SOTA methods on real-world SISR. The best and second best results are highlighted in \textbf{bold} and \underline{underline}.}
  \vspace{-1mm}
  \centering
  \setlength\tabcolsep{2pt}
  \begin{tabular}{lcccccc}
  \hline
  \multicolumn{1}{l}{Methods} & MUSIQ$\uparrow$ & LIQE$\uparrow$  & NRQM$\uparrow$  & NIQE$\downarrow$   & PI$\downarrow$    \\ \hline
  FlowIE~\cite{FlowIE}                  & 56.83  &2.44  & 4.84 & \underline{5.68}  & 5.53 \\
  ResShift~\cite{Resshift}     & 56.14  &2.80 & 6.20 & 7.34  & 5.55 \\
  SinSR~\cite{SinSR}     & \underline{61.45}  &\underline{3.19}  & \textbf{6.72} & 5.76   & \underline{4.49} \\ \hline
  Ours      & \textbf{64.04} & \textbf{3.79}  & \underline{6.70} & \textbf{3.55}  & \textbf{3.49} \\ \hline
  \end{tabular}
  \label{tabl_compare_realsr}
\end{table}
\begin{figure}[!h]
  \vspace{-2mm}  
  \centering
  \includegraphics[width=0.95\columnwidth]{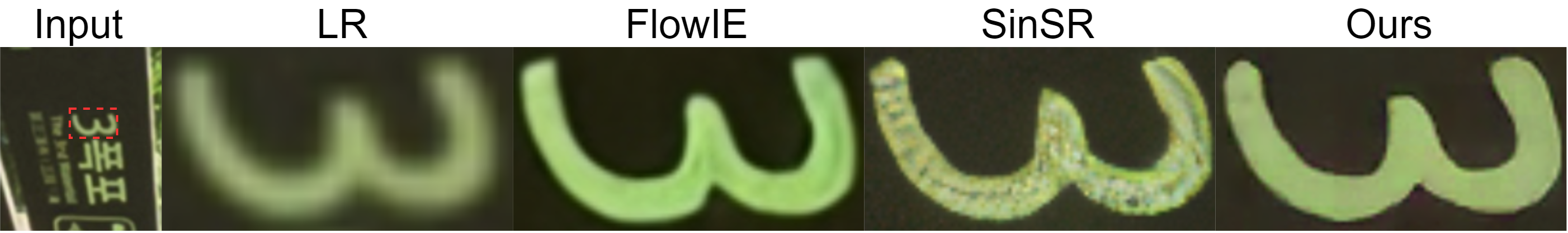} 
  \vspace{-2mm}
  \caption{Qualitative comparison on RealSR.}
  \vspace{-5mm}
  \label{fig_real_com1}
\end{figure}
\noindent{\textbf {Real-world image restoration.}} \ \
We further explore the capability of DTWSR on real-world image restoration task. We train DTWSR on ImageNet training set~\cite{ImageNet} following the pipeline in ResShift~\cite{Resshift} with the degradation model from RealESRGAN~\cite{RealESRGAN} adopted, and evaluate it on RealSR data\cite{realsr}. A series of non-reference metrics, \eg., MUSIQ~\cite{MUSIQ}, LIQE~\cite{LIQE}, NRQM~\cite{NRQM}, NIQE~\cite{NIQE} and PI~\cite{PI} are employed to justify the restoration quality, following common practice. As shown in Table \ref{tabl_compare_realsr}, our method shows promising quality, surpassing existing methods on most metrics. As shown in Fig.~\ref{fig_real_com1}, our method produces more natural results with clearer edges. FlowIE exhibits noticeable color drift, and SinSR introduces excessive noise. 


\begin{table*}[]
  \caption{Ablations of our approach for $8\times$ face SISR. Both components can introduce positive impact, while their fusion combines the strength of both components, resulting in the best scores.}
  \vspace{-2mm}
  \setlength\tabcolsep{3pt}
  \begin{tabular}{lccccccccccc}
  \hline
  \multirow{2}{*}{Model} & \multirow{2}{*}{\begin{tabular}[c]{@{}c@{}}SISR on \\ spectra\end{tabular}} & \multicolumn{4}{c}{WSDT}                                                                                                                                                                       & \multirow{2}{*}{Tokens$\downarrow$} & \multirow{2}{*}{PSNR$\uparrow$} & \multirow{2}{*}{SSIM$\uparrow$} & \multirow{2}{*}{Cons.$\downarrow$} & \multirow{2}{*}{Deg.$\downarrow$} & \multirow{2}{*}{FID$\downarrow$} \\ \cline{3-6}
                         &                                                                             & \begin{tabular}[c]{@{}c@{}}Pyramid\\ tokenization\end{tabular} & \begin{tabular}[c]{@{}c@{}}Dual \\ decoder\end{tabular} & \begin{tabular}[c]{@{}c@{}}LF \\ residual\end{tabular} & \begin{tabular}[c]{@{}c@{}}Attention \\ mask\end{tabular} &                         &                       &                       &                        &                       &                      \\ \hline
  Pixel-DiT              & -                                                                        & -                                                           & -                                                    & -                                                   & -                                                      & 1040                    & 23.94                 & 0.714                 & 1.77                   & 53.47                 & 61.372               \\
  Freq-DiT               & $\surd $                                                                         & $\surd $                                                            & -                                                    & -                                                   & -                                                      & 704                     & 24.02                 & 0.715                 & 0.582                  & 55.17                 & 60.352               \\
  DTWSR(a) & $\surd $                                                                         & $\surd $                                                            & $\surd $                                                     & -                                                   & -                                                      & 704                     & 24.06                 & 0.719                 & 2.35                   & 54.40                 & 58.531               \\
  DTWSR(b)              & $\surd $                                                                         & $\surd $                                                            & $\surd $                                                     & $\surd $                                                    & -                                                      & 704                     & 24.09                 & 0.718                 & 0.549                  & 56.19                 & 60.008               \\
  DTWSR(c)              & $\surd $                                                                         & -                                                           & $\surd $                                                     & $\surd $                                                    & $\surd $                                                       & 1040                    & 23.63                 & 0.694                 & 1.147                  & 57.30                 & 66.051               \\
  DTWSR(ours)            & $\surd $                                                                         & $\surd $                                                            & $\surd $                                                     & $\surd $                                                    & $\surd $                                                       & 704                     & 24.09                 & 0.719                 & 0.503                  & 53.85                 & 56.771               \\ \hline
  \end{tabular}
  \vspace{-3mm}
  \label{tab_abls}
\end{table*}

\begin{table}[]
  \caption{Ablation on the effect of interrelations among multi-scale frequency sub-bands.}
  \centering
  \vspace{-2mm}
  \begin{tabular}{lccccc}
  \hline
  & PSNR$\uparrow$  & SSIM$\uparrow$  & Cons.$\downarrow$ & Deg.$\downarrow$  & FID$\downarrow$    \\ \hline
  w/o & 23.99 & 0.711 & 1.154 & 55.00 & 61.670 \\
  w   & 24.09 & 0.719 & 0.503 & 53.85 & 56.771 \\ \hline
  \end{tabular}
  \label{tab_abl_wo_relation}
  \vspace{-3mm}
\end{table}
\subsection{Ablation studies}
We conduct ablation studies on $8\times$ face SISR to evaluate the effectiveness of SISR on wavelet spectra and the proposed denoising network WSDT.

\noindent{\textbf {Effects of SISR on frequency domain. }} 
To show the effect of SR on wavelet frequency domain, we train a diffusion transformer model on pixel domain, named Pixel-DiT. Images are patchified spatially using a patch size of $4$~\cite{VIT} and LR tokens are concatenated to provide in-context conditioning. For fair comparison, we use the reconstruction loss as DTWSR with constraints on both pixel and frequency domains. As shown in Table \ref{tab_abls}, compared to Pixel-DiT, DTWSR enhances SR performance using only two-thirds the number of tokens.

\noindent{\textbf {Ablation on WSDT architecure. }}
We perform ablation study on WSDT architecture, including the dual decoder design, LF Residual, attention mask and pyramid tokenization. We define Freq-DiT as our basic denoising network. It applies diffusion transformer on wavelet spectra but uses a unified decoder for all frequency sub-bands. In contrast, DTWSR(a) decodes LF and HF sub-bands independently by two decoders, without LF Residual considered.  DTWSR(b) takes LF Residual into account but does not think over its influence on HF tokens. DTWSR(c) employs equal patch size ($4$) across frequency sub-bands. 

As illustrated in Table \ref{tab_abls}, the distinct distribution in LF and HF affects each other in Freq-DiT, leading to worse results on Deg. and perceptual quality FID. When isolating decoders, DTWSR(a) shows better FID score straightaway. However, without aligning the relations between LF and HF sub-bands by LF Residual, it shows poor Cons.. DTWSR(b) Introduces LF Residual to promotes the alignment between LF and HFs, and shows improved Cons.. However, it re-suffers the influence from LF to HF, leading to worse FID and Deg.. Therefore, we design attention mask $M_{high}$ further in DTWSR(ours) to force the re-alignment between LF and HF sub-bands but avoid overwhelming of HF, leading to the best performance on both fidelity and perceptual quality. 


Compared to DTWSR(c), DTWSR(ours) uses much less tokens, but achieves much better generation quality, which prove the advantage of our pyramid tokenization method.  

\noindent{\textbf {Effects of correlation among frequencies. }} 
We employ a designed attention mask to artificially remove the interaction between different levels of HF sub-bands. As shown in Table \ref{tab_abl_wo_relation}, when the model ignores the interrelations among HF sub-bands, performance decreases across all metrics. The drop is particularly notable in FID, indicating significant degradation in image textural details.

\vspace{2mm}
\section{Conclusion}
In this work, we propose a diffusion transformer model based on image multi-level wavelet spectra, offering a novel solution for SISR. Our method integrates the strengths of diffusion models and transformers to capture the complex interrelations among multi-scale HF sub-bands. The pyramid tokenization promotes the learning of relationships between sub-bands for transformer. A dual-decoder transformer model is designed which separately processes the smooth contents in LF and the sparse HF details. The dedicated designed HDDec  facilitates the exploration of correlations among frequency sub-bands, resulting in SR images with improved objective fidelity and perceptual quality. Extensive experiments demonstrate that our method achieves state-of-the-art performance on SISR across various SR magnification and diverse datasets. In the future, we intent to further explore the potential of multi-level wavelet spectra in promoting image generation.

\clearpage
{
    \small
    \bibliographystyle{ieeenat_fullname}
    \bibliography{main}
}

\end{document}